\pdfoutput=1
\documentclass[11pt]{article}
\usepackage[preprint]{acl}
\usepackage{times}
\usepackage{latexsym}
\usepackage[T1]{fontenc}
\usepackage[utf8]{inputenc}
\usepackage{microtype}
\usepackage{inconsolata}
\usepackage{graphicx}
\usepackage{booktabs}
\usepackage{amsmath,amssymb}
\usepackage{tikz}
\usepackage{placeins}
\usepackage{float}
\usetikzlibrary{positioning,arrows.meta}

\title{Better Deck or Different Judge? Evaluating Agentic Harness Gains in Corporate and Investment Banking}

\author{Ludovic Gibert \\
  AIData2Action \\
  {\small\texttt{ludovic.gibert@protonmail.com}} \\\And
  Matis Despujols \\
  TW3 Partners \\
  {\small\texttt{m.despujols@tw3partners.com}} \\\And
  Andr\'{e}-Louis Rochet \\
  TW3 Partners \\
  {\small\texttt{arochet@tw3partners.com}} \\}

\begin{document}
\maketitle

% NUMBERS-FROM: data/counts.json; data/textonly.json (blind); data/audit_checks.json.
\begin{abstract}
Corporate and investment banking teams use presentations to support credit decisions and advise clients on financing and transactions. Producing these decks requires reconciling financial data, tracing sources and turning analysis into a recommendation. We retrospectively study the development of an agentic harness combining a 27B language model, financial calculations, narrative templates and validation checks. LLM judges guide engineering changes and assess the resulting decks, raising the question of whether higher scores reflect better documents or changes in grading. In shared-session text-only grading with template markers removed, five judges score the complete system 20.4 to 33.6 points out of 95 above the same model generating directly from a short prompt. Every judge scores the system higher on all seventeen development deliverables. Margins against direct Opus generation from a short prompt range from $-4.7$ to $+0.8$ points. Judges agree on broad progress across development rounds but agree less on final-deck rankings than on pooled scores. Repeated grading also shifts scores on unchanged decks, making small improvements difficult to distinguish from judge variability.
\end{abstract}

\section{Introduction}\label{sec:intro}

Corporate and investment banking (CIB) presentations turn financial analysis into decisions about credit, financing and transactions. A recommendation must be supported by figures that agree across slides and can be traced to sources. Automating these documents therefore involves more than drafting prose.

During development, an LLM judge can apply an expert rubric \citep{zheng2023judging,liu2023geval} and return a score, a verdict and a list of defects. In our development loop, these assessments guided changes to prompts, templates and code. The judge served both as a source of feedback and as the measure of whether those changes helped.

Scores also move when the documents have not changed. The same judge can grade the same deck differently on a second pass, and two configurations of the same judge can apply the rubric at different levels. Prior work documents position, verbosity and self-preference biases \citep{wang2024fair,panickssery2024self,ye2024justice} and the self-inconsistency of repeated LLM ratings on standard benchmarks \citep{haldar2025rating}. Development teams, however, decide in rubric points and in a ship-or-not verdict, and reliability is rarely reported in those units for long documents.

% NUMBERS-FROM: data/counts.json; data/audit_checks.json.
We measured this during the development of a CIB presentation system (Section~\ref{sec:harness}). Seventeen deliverables were graded over eight development rounds. A later panel of eight judges from six model families compared development versions, direct-generation baselines and two text-only passes, the second with template markers removed. We examine score repeatability, agreement on progress and final rankings, and the advantage of the complete system over direct generation.

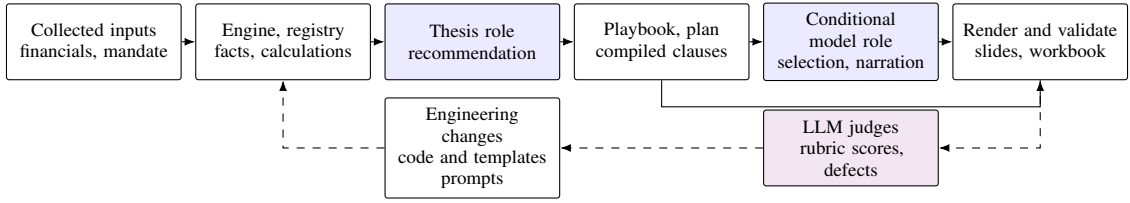
\begin{figure*}[t]
\centering
\begin{tikzpicture}[font=\scriptsize,
  box/.style={draw, rounded corners=1pt, align=center, minimum height=10mm, text width=2.05cm, fill=white},
  llm/.style={box, fill=blue!8}, jud/.style={box, fill=violet!10},
  arr/.style={-{Latex[length=1.4mm]}}, loop/.style={-{Latex[length=1.4mm]}, dashed}]
\foreach \n/\x/\st/\t in {inp/0/box/{Collected inputs\\ financials, mandate},
                          reg/2.5/box/{Engine, registry\\ facts, calculations},
                          th/5/llm/{Thesis role\\ recommendation},
                          pl/7.5/box/{Playbook, plan\\ compiled clauses},
                          wr/10/llm/{Conditional model role\\ selection, narration},
                          ren/12.5/box/{Render and validate\\ slides, workbook}}
  \node[\st] (\n) at (\x,0) {\t};
\foreach \u/\v in {inp/reg, reg/th, th/pl, pl/wr, wr/ren} \draw[arr] (\u) -- (\v);
\draw[arr] (pl.south) -- ++(0,-0.35) -| (ren.south);
\node[jud] (jd) at (10,-1.4) {LLM judges\\ rubric scores, defects};
\node[box] (dev) at (5,-1.4) {Engineering changes\\ code and templates\\ prompts};
\draw[loop] (ren.south) |- (jd.east);
\draw[loop] (jd) -- (dev);
\draw[loop] (dev.west) -| (reg.south);
\end{tikzpicture}
\caption{The hybrid generation system and its development loop. Blue boxes denote model roles; deterministic prose can bypass the conditional model role. The evaluated contrast starts from collected inputs. Collection services and the application's banker and supervisor review loop lie outside this contrast.}
\label{fig:harness}
\end{figure*}

\section{Related work}\label{sec:related}

Rubric-guided judges can correlate with human ratings \citep{zheng2023judging,liu2023geval,chiang2023can}, with documented position, length and self-preference biases \citep{wang2024fair,dubois2024length,panickssery2024self,ye2024justice}. Agreement depends on the task \citep{bavaresco2024llms,thakur2024judging}. Surveys, trained judges and adversarial benchmarks address accuracy against references \citep{gu2024survey,kim2024prometheus,zhu2023judgelm,zeng2024llmbar}.

\citet{haldar2025rating} study repeated-rating reliability, \citet{schroeder2024trust} treat judgments as draws from a distribution, and \citet{atil2024nondeterminism} document variation under nominally deterministic settings. \citet{li2026whodrifted} distinguishes system drift from judge drift using human anchors and sequential inference. Our retrospective study measures repeated scores during presentation development, without human anchors or a drift-detection guarantee.

Clustered inference matters when evaluation items share context \citep{miller2024errorbars}; seed variation and power also constrain comparisons \citep{madaan2024variance,card2020power,dror2018hitchhiker}. We use established agreement statistics \citep{shrout1979intraclass,mcgraw1996forming,koo2016guideline}, detectable-change estimates \citep{weir2005quantifying}, kappa \citep{cohen1960coefficient,cohen1968weighted}, paired differences \citep{bland1986statistical} and rank tests \citep{wilcoxon1945individual}.

Financial benchmarks span numerical reasoning and question answering \citep{chen2021finqa,islam2023financebench,xie2024finben}. MBABench evaluates complete financial spreadsheets with expert validation \citep{yen2026mbabench}. Slide-generation research includes document summarisation, agentic generation and broader PowerPoint evaluation \citep{fu2022doc2ppt,zheng2025pptagent,gandhi2026ppteval}. Harness evaluation also requires matched feedback and compute budgets and reserved tasks \citep{wang2026harnessevolution}. We study repeated and cross-judge assessments of complete CIB decks, comparing an iteratively developed harness with direct generation.

\section{The generation harness}\label{sec:harness}

% NUMBERS-FROM: data/provenance_audit.json (architecture); data/panel.csv (model identifiers).
The system combines an on-premises Qwen3.8-27B model with financial calculations, playbooks and output checks (Figure~\ref{fig:harness}). It builds on a staged pipeline for debt capital markets and coverage pitches (Appendix~\ref{app:pipeline}). We evaluate the resulting hybrid system, with explicit separation between model roles and compiled prose.

The evaluated contrast starts from collected financial inputs and a mandate. A deterministic engine constructs a typed registry of facts, with value, unit, period and provenance class. Sources and formulas are recorded where applicable. Unavailable bank-held information is represented as a named gap. Collection services exist in the wider application but are not compared here.

Playbooks determine the page plan. A model role states a recommendation under the input constraints. Narrative production combines clauses compiled from registry facts, explicit prose templates and conditional model selection, ordering or writing. Template pages can bypass the writer call. Output validation rejects unsupported numeric literals after generation; this is not constrained decoding. Editing and rendering then produce the presentation and a controls workbook.

Historical templates contain long literal passages found in every evaluated final deck. However, the archives do not link each deck to a complete code, input and generation-log snapshot. We therefore cannot quantify the model-written share of the evaluated prose. Later logs with fewer writer calls cannot establish that share retrospectively. Figure~\ref{fig:example} illustrates the registry mechanism.

After rendering, checks cover model configuration, registry consistency, narrative contradictions and layout defects. These checks determine whether the generated files pass validation. The application's supervisor and banker review loop is outside this evaluation.

% NUMBERS-FROM: data/illustrative_example.json (explicitly invented, not empirical).
\begin{figure}[t]
\centering\footnotesize
\fbox{\begin{minipage}{0.94\columnwidth}\raggedright
Registry entry. \texttt{LEV-NET-PF} = 2.99x; period FY2026 pro forma; class computed; formula \texttt{DEBT-NET-PF / EBITDA-FY26}; sources: annual report 2025, mandate.\\[3pt]
Narrative slot. ``The recommended sequence brings pro forma net debt to EBITDA to \{\{LEV-NET-PF\}\}, below the agency threshold of \{\{THR-AGENCY\}\}.''\\[3pt]
Rendered. ``\ldots to 2.99x, below the agency threshold of 3.5x.''\\[3pt]
Refused. ``\ldots to about 3.0x'' contains a digit outside a key and fails output validation. A title claiming a fall contradicts the registered increase from 2.53x to 2.99x and fails the narrative-consistency gate.
\end{minipage}}
\caption{An invented illustration of registry-based rendering and validation, translated from French. It is not an empirical test of the checks.}
\label{fig:example}
\end{figure}

\section{Setting and protocol}\label{sec:setting}

The seventeen deliverables span coverage documents, credit reviews, financing pitches and M\&A advisory (Table~\ref{tab:decks}). Each is an exercise on a listed or rated issuer built from public information. Between rounds, coding agents changed templates, the engine and the prompts, mostly to fix defects quoted in the previous round's verdicts and partly on written feedback from a practitioner.

\begin{table}[t]
\centering\scriptsize\setlength{\tabcolsep}{3pt}
\begin{tabular}{lll}
\toprule
Session & Deliverable & Sector \\
\midrule
Coverage (3) & Company credit fact sheet & Bus.\ services \\
 & Coverage summary & TMT \\
 & Teaser and info.\ memorandum & TMT \\
Credit (5) & LBO participation review & Industrials \\
 & Project-finance review & Energy \\
 & Credit committee, green loan & Technology \\
 & Annual credit review & Bus.\ services \\
 & Annual credit review & TMT \\
Financing (5) & Share buyback programme & Technology \\
 & Refinancing of acquisition debt & TMT \\
 & Rating advisory & TMT \\
 & Sustainability-linked bond & Industrials \\
 & Green financing pitch & Technology \\
M\&A (4) & Acquisition bond financing & Bus.\ services \\
 & Offer price assessment & Consumer \\
 & Sell-side floor price & Technology \\
 & IPO preparation & Technology \\
\bottomrule
\end{tabular}
\caption{The seventeen deliverables. One judge session grades one group of three to five decks.}
\label{tab:decks}
\end{table}

% NUMBERS-FROM: data/panel.csv (criterion scores and observed maxima, checked against the rubric).
The rubric follows senior-banker review practice. Reader orientation, ten-second synthesis, recommendation and financial depth each receive 15 points. Consistency and auditability receive 20, title storyline 10, why this bank 5 and visual quality 5. For each deck it writes criterion scores with quotations, a total, a verdict (not sendable, after one round of corrections, send as is) and the five worst defects (Appendix~\ref{app:instruction}). Decks receive fresh random letters in every campaign, and the judge sees the rendered slides and the extracted text with no information about authorship or round.

% NUMBERS-FROM: data/panel.csv (judge, round, date, pass and configuration).
For agent-based judges, each session is fresh, with file-reading tools, no memory of other sessions and default sampling settings (Appendix~\ref{app:config}). The panel used Claude Opus 5.5, Sonnet 5.5 and Haiku 4.5; the development campaigns used the Opus and Sonnet versions served at the time, which we could not pin. The development campaigns (24 and 25 September 2026) used Opus, plus Sonnet in rounds 5 and 6, in a configuration we cannot reconstruct exactly. The panel (29 September) used a controlled configuration with no additional system instructions and tools restricted to reading and writing files; it graded rounds 1, 5 and 7 with Sonnet and rounds 1 and 7 with Haiku. An earlier panel pass of Sonnet and Haiku on rounds 1 and 7 ran with an extraneous system instruction unrelated to the task, and we use it only as a second pass.

% NUMBERS-FROM: data/counts.json (configuration checks); data/panel.csv (judge configurations).
The cross-family panel graded the round-1 and round-7 decks with DeepSeek V4.1 Flash, GLM 5.3 FlashX, GPT-6 Luna Pro and Gemini 3.8 Flash through their commercial APIs, and with Qwen3.8-27B on the on-premises GPU server that serves the generator. Each session is one request carrying every slide image, downscaled to 1600 pixels, and the extracted text, with the same French instruction and one added paragraph requesting a closing JSON block of scores. Commercial API judges use low reasoning effort; the on-premises judge has no effort setting. We log the serving provider of every call and, for the API judges, reasoning tokens and deck position. DeepSeek, GLM, GPT-6 Luna Pro and Qwen graded each round twice, once in the original sessions and once with decks reassigned to four random mixed sessions. As a check on the configuration, Sonnet graded round 7 once more with the downscaled images and the JSON paragraph and moved by 1.4 points. Opus graded round 7 and the no-harness sets in the controlled configuration, 1.1 points above its development-campaign score on round 7, and Table~\ref{tab:harness} uses these configuration-matched scores.

% NUMBERS-FROM: data/panel.csv (conditions); data/provenance_audit.json (input provenance).
For the no-harness condition, Qwen3.8-27B (on premises) and Opus each wrote the seventeen deliverables directly. They received collected financials, a mandate, decisions to make, a one-line deliverable description and a request for 10 to 18 slides as JSON. The archived Opus prompts match the current input files, but the historical harness inputs lack snapshots to verify exact equality. A neutral template rendered the slides in the same image and text format as the harness decks. The condition omits the fact engine, playbooks, compiled narration and gates, while retaining collected inputs. For a stronger baseline, both models also received a longer prompt that states the rubric, the conventions of a CIB pitch book and a self-check before answering. The conventions cover action titles, a synthesis page, a dated ask, scenarios, named data gaps and figures reconciled across pages. The 27B decks for this prompt were generated through the API, eight with medium and nine with low reasoning effort. The strong-prompt 27B decks were graded in mixed sessions that also contain the harness decks of the same deliverables, which removes the between-session offset for that comparison. These decks also differ from the short-prompt ones in serving (the API instead of the on-premises server) and reasoning effort, not only in the prompt. The seven judges other than Haiku graded both short-prompt sets in the original sessions; Opus, Sonnet, DeepSeek and GLM also graded the strong-prompt Opus decks, in separate sessions. The rubric and instruction were otherwise identical in every campaign except the text-only passes described below.

% NUMBERS-FROM: data/textonly.json; data/counts.json.
Two text-only passes remove the slides. Each original session holds three versions per deliverable under fresh letters and in random order, giving 9 to 15 texts. The second pass applies identical normalisation rules to every version. It removes agenda pages, confidentiality footers, page numbers, section codes, capitalised headings and bullet glyphs, and standardises table rows and slide breaks. Wording, structure and length still differ. The instruction sets visual quality to zero, giving a total out of 95, and asks judges not to penalise absent layout elsewhere. Opus, Sonnet, DeepSeek, GLM and GPT-6 Luna Pro each graded both passes once. We use the second pass as the main text-only result. The first retained identifying template material and is summarised in Appendix~\ref{app:crit}.

Rounds 7a and 7b are two Opus passes on byte-identical text and images, run concurrently, as are rounds 8a and 8b with a controls workbook added. Round 5 was graded by Opus, by Sonnet in the development campaign, and by Sonnet again in the panel, on the same files.

% NUMBERS-FROM: data/results.json (retest and pooled_retest); data/audit_checks.json.
For two passes of a judge on the same $n$ documents, with per-document differences $d_i$, a conventional per-document minimal detectable change estimate is
\begin{equation}
\mathrm{MDC}_{95} = 1.96\,\sqrt{2}\,\mathrm{SEM} = 1.96\,\mathrm{sd}(d)
\end{equation}
for one document \citep{weir2005quantifying}. For a mean over $n$ documents with independent errors it is $1.96\,\mathrm{sd}(d)/\sqrt{n}$. Intervals use 4,000 bootstrap resamples of decks, or a chi-square calculation for $\mathrm{sd}(d)$. Deck-level intervals and Wilcoxon and sign-flip tests assume independence across decks. We report them as exploratory and add sign-flip sensitivities at the session level. These also assume symmetric differences and have limited resolution with four sessions.

% NUMBERS-FROM: data/counts.json; data/audit_checks.json.
We preserve fractional scores and use closing JSON totals when they differ from narrative prose, logging each discrepancy. Analyses omit missing scores. Appendix~\ref{app:archive} gives the evaluation counts, missing values and extraction checks.

\section{Results}\label{sec:results}

\subsection{Two passes of the same judge}\label{sec:res-retest}

% NUMBERS-FROM: data/results.json (retest_r7, retest_r8, pooled_retest, robust).
Two Opus passes on identical decks agree at an ICC of 0.77 in round 7 and 0.78 in round 8 (Table~\ref{tab:reliability}). The point estimates are in the ``good'' band of \citet{koo2016guideline}, but their intervals reach down to 0.31 and 0.44. Pooling the 34 pairs gives $\mathrm{sd}(d)=4.05$ (interval 3.3 to 5.3), yielding a per-deck noise scale of 7.9 points (6.4 to 10.4), or 8.4 when the systematic shift is included. These estimates depend on the independence and distributional assumptions in Section~\ref{sec:setting}.

\begin{table*}[t]
\centering\small\setlength{\tabcolsep}{5pt}
\begin{tabular}{lccc}
\toprule
 & Opus / Opus, round 7 & Opus / Opus, round 8 & Opus / Sonnet, round 5 \\
\midrule
Campaign means & 75.8 / 77.8 & 75.7 / 76.8 & 75.0 / 86.8 \\
Mean shift [95\% CI] & +2.0 [0.1, 3.8] & +1.1 [$-$0.8, 2.9] & +11.8 [8.7, 14.6] \\
Wilcoxon $p$ / permutation $p$ & 0.033 / 0.063 & 0.31 / 0.31 & $<$0.001 / $<$0.001 \\
Shift by session (Cov, Cre, Fin, M\&A) & +0.7, +5.6, +1.2, $-$0.5 & +4.7, +0.2, +1.0, $-$0.3 & +16.0, +6.0, +14.8, +12.0 \\
sd($d$), MDC$_{95}$ per deck & 3.97, 7.8 & 4.20, 8.2 & 6.28, 12.3 \\
Spearman $\rho$ & 0.71 & 0.69 & 0.43 \\
ICC(A,1) [95\% CI] & 0.77 [0.31, 0.92] & 0.78 [0.44, 0.89] & 0.18 [0.00, 0.32] \\
Verdicts equal, Cohen's $\kappa$ & 12/17, 0.11 & 15/17, 0.60 & 6/17, $-$0.19 \\
``Send as is'' verdicts & 0 / 0 & 0 / 0 & 0 / 5 \\
\bottomrule
\end{tabular}
\caption{Agreement between two judge passes on identical decks ($n=17$ per column). $d$ is the per-deck difference in total, second pass minus first. ICC(A,1) is the absolute-agreement, single-rater coefficient \citep{mcgraw1996forming}; intervals are bootstrap over decks. The Sonnet column is the development pass of 24 September.}
\label{tab:reliability}
\end{table*}

\subsection{Shifts cluster by judge session}\label{sec:res-session}

% NUMBERS-FROM: data/results.json; data/audit_checks.json.
A single session carried most of each replication shift, the credit decks in round 7 and the coverage decks in round 8 (Table~\ref{tab:reliability}). Decks graded together appear to share an offset. For unchanged round-7 decks, the Wilcoxon test gives $p=0.033$, the deck-level sign-flip test $p=0.063$ and its session-level counterpart $p=0.375$. The root mean square of the two campaign shifts, multiplied by 1.96, is 3.2 points, compared with 1.9 under independent deck errors. With only two shifts, 3.2 is an exploratory noise scale, not a calibrated decision threshold.

\begin{table*}[t]
\centering\small\setlength{\tabcolsep}{5pt}
\begin{tabular}{lccccc}
\toprule
 & Harness & \multicolumn{2}{c}{27B without harness} & \multicolumn{2}{c}{Opus without harness} \\
\cmidrule(lr){3-4}\cmidrule(lr){5-6}
Judge & (27B, round 7) & short prompt & strong prompt$^{a}$ & short prompt & strong prompt \\
\midrule
Claude Opus & 76.9 & +16.9 (17/17) & +20.4 (17/17) & $-$2.1 (8/17) & $-$3.4 (6/17) \\
Claude Sonnet & 75.5 & +13.6 (15/17) & +19.1 (16/17) & +1.1 (12/17) & $-$1.4 (8/17) \\
DeepSeek V4.1 Flash & 89.1 & +12.9 (17/17) & +17.1 (16/17) & +3.2 (12/17) & +1.6 (10/16) \\
Qwen3.8-27B & 93.4 & +11.6 (16/17) &  & +4.3 (14/17) &  \\
Gemini 3.8 Flash & 91.5 & +10.7 (15/17) &  & +1.9 (10/17) &  \\
GLM 5.3 FlashX & 87.8 & +10.5 (17/17) & +14.2 (17/17) & +2.3 (12/17) & $-$0.5 (6/17) \\
GPT-6 Luna Pro & 79.8 & +8.5 (15/17) & +10.6 (14/16) & +4.2 (12/17) &  \\
\bottomrule
\end{tabular}

\caption{Advantage of the harness deck (27B, round 7) over each no-harness condition, in points of the mean total, with the number of deliverables where the harness deck scores higher. $^{a}$Graded in mixed sessions that also contain the harness decks; the difference is taken against the harness decks of those same sessions. Opus is graded here in the panel configuration; a blank cell means the judge did not grade that condition, and a count out of 16 means one deck's score is missing.}
\label{tab:harness}
\end{table*}

\begin{figure}[t]
\centering
\includegraphics[width=\columnwidth]{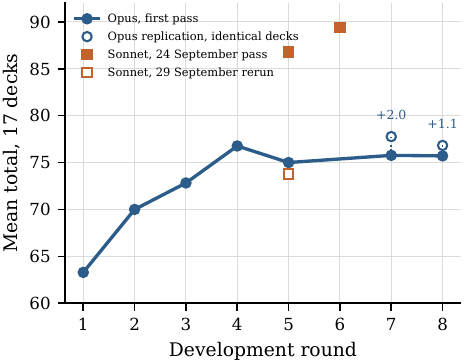}
\caption{Opus campaign mean over the 17 decks by round, with Sonnet passes marked. Open points repeat assessments on identical decks; labels show the observed mean shifts.}
\label{fig:trajectory}
\end{figure}

\begin{table}[t]
\centering\scriptsize\setlength{\tabcolsep}{4pt}
\begin{tabular}{lrrrr}
\toprule
Rounds & $\Delta$ mean & Up & Down & Wilcoxon $p$ \\
\midrule
1 $\to$ 2 & +6.7 & 10 & 7 & 0.169 \\
2 $\to$ 3 & +2.8 & 15 & 1 & 0.003 \\
3 $\to$ 4 & +3.9 & 14 & 3 & 0.003 \\
4 $\to$ 5 & $-$1.8 & 4 & 11 & 0.064 \\
5 $\to$ 7 & +0.8 & 9 & 8 & 0.585 \\
7 $\to$ 8 & $-$0.1 & 8 & 6 & 0.975 \\
\bottomrule
\end{tabular}

\caption{Change in the Opus campaign mean between consecutive rounds (first passes), with decks up and down, alongside the exploratory replication scale of 3.2 points (Section~\ref{sec:res-session}; replications in Table~\ref{tab:reliability}). Round 6 has no Opus pass; values are rounded independently.}
\label{tab:transitions}
\end{table}

\subsection{An offset that did not replicate}\label{sec:res-offset}

% NUMBERS-FROM: data/results.json (model_r5).
Sonnet's development pass scored the round-5 decks 11.8 points above Opus, with every deck at or above Opus. It sent five decks as is that Opus returned for corrections (Figure~\ref{fig:pairs} in Appendix~\ref{app:pairs}). Taken at face value, this offset is larger than any Opus transition in the loop.

% NUMBERS-FROM: data/panel_results.json (sonnet_r5_rerun, per_judge); data/results.json.
Five days later, the panel's Sonnet rerun of the same round-5 files averaged 73.8, with no ``send as is'' verdict. That is a difference of $-1.2$ points from Opus (interval $-3.8$ to $+1.2$) and of $-13.0$ points from the earlier Sonnet pass ($-14.9$ to $-11.1$), for a cause we cannot identify. Settings, tool access and the served model version may all have differed. The pass with an extraneous system instruction differed by $-0.4$ points; that comparison does not isolate the instruction effect. The rerun still ranks the decks differently from Opus ($\rho=0.55$, $\mathrm{sd}(d)=5.4$). A model name was not enough to reproduce that judge's level, and round 6, graded only by the earlier Sonnet setup at 89.4, cannot be compared with any Opus round.

\begin{figure*}[t]
\centering
\includegraphics[width=0.92\textwidth]{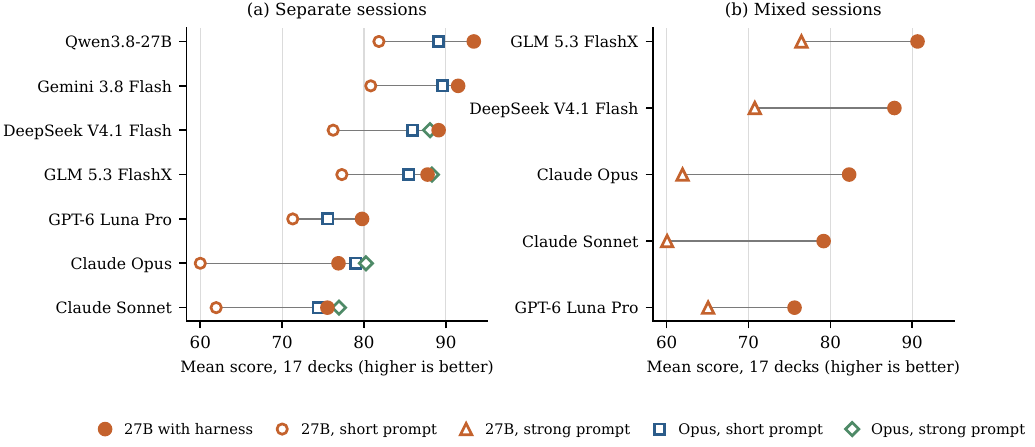}
\caption{Mean score per judge for the 27B with the harness (filled) and for decks written without it (open). Panel (a) compares separate-session means for the harness, the direct 27B with a short prompt, and direct Opus with a short or strong prompt. Panel (b) compares the strong-prompt 27B and harness decks graded together in mixed sessions. Paired differences and deck counts are in Table~\ref{tab:harness}.}
\label{fig:harness-effect}
\end{figure*}

\subsection{Eight judges from six model families}\label{sec:res-panel}

Every judge sees the decks improve between round 1 and round 7 (Table~\ref{tab:crossfam}). The gain ranges from 5.6 points (Haiku) to 13.9 points (DeepSeek), covers 11 to 14 of the 17 decks, and passes a Wilcoxon test at $p<0.03$ for every judge except Haiku ($p=0.11$). Judges that did not steer the loop are the useful check (Section~\ref{sec:threats}), and those from five other families see gains of 7 to 14 points.

The same round-7 decks average 75.5 with Sonnet and 93.4 with Qwen3.8-27B, and the lenient judges (Qwen, GLM, Gemini) send 9 or 10 of the 17 decks as is where Opus, Sonnet and GPT-6 Luna Pro send none. A descriptive generalizability analysis (Appendix~\ref{app:variance}) of the round-7 scores attributes 59\% of the variance to the judge, 14\% to the deck and 27\% to their interaction and error. In round 1, when the decks still differed widely, the deck carried 59\% and the judge 23\%. Removing Haiku changes these shares by at most 4.4 points. The eight judges were chosen by convenience, so the shares describe this panel and do not generalise to other judges. Averaging judges helps rankings more than levels (Figure~\ref{fig:dstudy} in Appendix~\ref{app:agree}). The relative coefficient reaches 0.72 with five judges, while the absolute one rises only to 0.45.

\begin{figure}[t]
\centering
\includegraphics[width=\columnwidth]{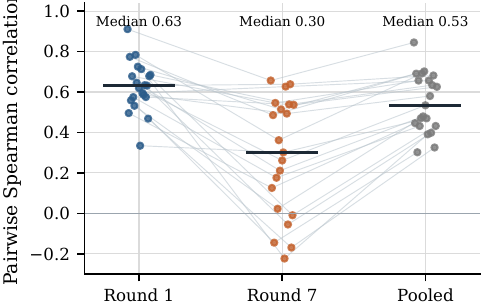}
\caption{Pairwise Spearman correlation among seven judges, excluding Haiku. Each line follows the same judge pair across round 1, round 7 and their pooled scores. Bars mark the median. Pooling includes the common improvement across rounds. Pairs share judges and decks and are not independent observations.}
\label{fig:round-agreement}
\end{figure}

% NUMBERS-FROM: data/audit_checks.json; data/crossfam.json.
Pooling rounds obscures weaker agreement on final rankings. Among the seven judges excluding Haiku, median pairwise rank correlation is 0.63 in round 1, 0.30 in round 7 and 0.53 when both rounds are pooled (Figure~\ref{fig:round-agreement}). Reduced dispersion among final decks may also lower correlation. Correlation with the pooled consensus ranges from 0.46 to 0.76. With decks reassigned to sessions, estimated per-deck detectable change is 13.5 to 16.1 points for the four judges measured that way (Table~\ref{tab:crossfam}).

\begin{table}[t]
\centering\scriptsize\setlength{\tabcolsep}{3pt}
\begin{tabular}{lrrrrrr}
\toprule
Judge & Round 1 & Round 7 & Gain & Up & $\rho_{\text{cons}}$ & MDC \\
\midrule
Claude Sonnet & 65.4 & 75.5 & +10.1 & 12 & 0.76 &  \\
GPT-6 Luna Pro & 71.8 & 79.8 & +8.0 & 14 & 0.72 & 15.4 \\
GLM 5.3 FlashX & 80.4 & 87.8 & +7.4 & 13 & 0.65 & 14.5 \\
Claude Opus & 63.3 & 75.8 & +12.5 & 13 & 0.61 & 7.8 \\
Gemini 3.8 Flash & 80.3 & 91.5 & +11.2 & 13 & 0.59 &  \\
Qwen3.8-27B & 84.8 & 93.4 & +8.6 & 11 & 0.59 & 13.5 \\
DeepSeek V4.1 Flash & 75.2 & 89.1 & +13.9 & 14 & 0.46 & 16.1 \\
Claude Haiku & 81.1 & 86.7 & +5.6 & 11 & 0.24 &  \\
\bottomrule
\end{tabular}

\caption{Mean total in rounds 1 and 7, gain, decks that improved, Spearman $\rho$ with the median of the other judges over the 34 deck versions, and per-deck MDC$_{95}$ between two passes. Opus values come from the development campaign, and its MDC from identical sessions; the other MDC values come from reshuffled sessions; a blank cell means no second pass (Gemini) or a second pass with an extraneous system instruction (Sonnet, Haiku).}
\label{tab:crossfam}
\end{table}

\subsection{Observed system comparisons}\label{sec:res-harness}

% NUMBERS-FROM: data/harness.json; data/audit_checks.json (headline, content_wins_range).
This comparison pairs iteratively developed harness outputs with single-sample direct generations on the development cases. Every judge that graded both conditions scores the harness decks above the same model's decks written from the short prompt (Table~\ref{tab:harness}, Figure~\ref{fig:harness-effect}). The gap ranges from 8.5 points (GPT-6 Luna Pro) to 16.9 points (Opus), the harness deck wins on 15 to 17 of the 17 deliverables, and every bootstrap interval excludes zero. Part of the gap sits in presentation, since the no-harness decks use a plain template. Leaving out visual quality and why this bank, the six remaining criteria (90 points) still show a gap of 6.7 to 14.1 points, with the harness deck ahead on 15 to 16 deliverables. Consistency and auditability show the largest content gap relative to criterion weight, 19\% against the short prompt and 32\% against the strong prompt. Financial depth follows against the short prompt (Table~\ref{tab:harness-crit}).

% NUMBERS-FROM: data/counts.json; data/harness.json; data/audit_checks.json (headline).
The stronger baseline does not close the observed gap. Compared across sessions, it moves the 27B by $-6.2$ to $+1.9$ points relative to the short prompt. In sessions that mix both conditions, the harness decks lead by 10.6 to 20.4 points with the five judges that graded them (8.6 to 18.1 on the six criteria) and are ahead on 14 to 17 deliverables. Mixing may add a contrast effect, since the harness decks themselves score between 4.1 points lower (GPT-6 Luna Pro) and 5.4 points higher (Opus) than in their round-7 sessions, so part of that range can come from the comparison itself.

% NUMBERS-FROM: data/harness.json; data/audit_checks.json (headline, mean_criterion_gap_fraction).
Against Opus writing without the harness, the 27B with the harness is close. With the short prompt the difference in total ranges from $-2.1$ (Opus judge) to $+4.3$ (Qwen judge), and with the strong prompt from $-3.4$ (Opus judge) to $+1.6$ (DeepSeek judge). The Opus judge gives the harness its smallest margin in both cases and prefers the Opus-written short-prompt content by 3.4 points on the six criteria (interval 0.2 to 6.9), which is consistent with the self-preference reported by \citet{panickssery2024self}. Against direct Opus with the short prompt, the six content gaps are close to zero. The remaining advantage lies mostly in visual quality (19\% of its weight) and, to a lesser extent, in why this bank. The Qwen judge's preference for the harness decks, produced by the system containing that model, is matched by GPT-6 Luna Pro and does not isolate self-preference.

% NUMBERS-FROM: data/textonly.json (blind); data/audit_checks.json.
The advantage persists after removing slides and template markers. Every judge scores the harness text above the direct 27B text on all seventeen deliverables, with mean gaps of 20.4 to 33.6 points out of 95 (Table~\ref{tab:textonly}). These shared-session gaps use a different scale from Table~\ref{tab:harness} and may include contrast effects.

% NUMBERS-FROM: data/textonly.json (blind); data/audit_checks.json.
Against direct Opus text, the margins range from $-4.7$ to $+0.8$. Only the Opus judge's deck-bootstrap interval excludes zero, favouring Opus text. Its nominal Wilcoxon $p=0.026$ becomes $0.25$ in the session-level sign-flip sensitivity. The minimum two-sided value with four sessions is $0.125$. Self-preference \citep{panickssery2024self} is therefore a possible explanation, not an identified effect. A non-significant difference also does not establish equivalence.

% NUMBERS-FROM: data/textonly.json (blind); data/audit_checks.json.
Median lengths are 4,996 words for the harness, 3,575 for Opus and 2,061 for the 27B. Correlations between length difference and the margin over Opus range from 0.02 to 0.54. Only the Opus correlation has nominal $p<0.05$, but the Sonnet estimate remains 0.46. Length-related bias remains possible \citep{dubois2024length}, including against the shorter 27B baseline. The second pass also changes random order and judge sampling, so the change between passes cannot isolate the effect of template removal.

\subsection{Progress across development rounds}\label{sec:res-progress}

% NUMBERS-FROM: data/results.json; data/audit_checks.json.
The Opus mean rose from 63.3 in round 1 to 76.8 in round 4 (Figure~\ref{fig:trajectory}, Table~\ref{tab:transitions}). Rounds 2 to 4 raised almost every deck; the first transition raised the mean by 6.7 points with ten decks up and seven down. Later campaign-mean transitions stayed within the exploratory 3.2-point scale, although individual decks sometimes improved by more than the per-deck noise estimate. Round 8 also added a controls workbook to the judge's input. These single-pass trajectories cannot establish whether later reconciliation work improved professional quality.

\subsection{Verdicts and arithmetic}\label{sec:res-verdict}

% NUMBERS-FROM: data/results.json; data/counts.json.
Opus never returned ``send as is'' in its 153 development records, and its effectively binary verdict tracks the score (ROC area 0.89). Seven of 34 repeated verdicts flipped, on decks scored between 66 and 80 (Appendix~\ref{app:verdict}). Opus made no addition errors in 357 complete records across development and panel conditions. Controlled Sonnet made one in 255, including the configuration check; development Sonnet made two in 34. The second Sonnet pass made one in 34, Haiku twelve in 31, GPT-6 Luna Pro none in 235 and Gemini one in 68. These counts compare criterion sums with stated totals after preserving fractional scores.

\section{Threats to validity}\label{sec:threats}

We compare judges with themselves and each other, without a banker-rated reference set. Agreement cannot establish accuracy or professional acceptability. The convenience panel includes related model families, and its variance shares and projected averaging benefits need not generalise. With one observation per deck and judge, the residual combines interaction and measurement error.

% NUMBERS-FROM: data/counts.json; data/audit_checks.json.
Seventeen deliverables in four sessions and two replication pairs constrain inference. Deck-level intervals ignore session dependence, while session-level sensitivities have little resolution. All significance tests are exploratory and unadjusted for multiple comparisons. More independently randomised passes and sessions would be needed for stable variance estimates. API and agent interfaces also differ. DeepSeek used six providers, including one serving 97 of its 255 records.

The harness comparison combines calculations, playbooks, compiled prose, model roles, validation and presentation. No matched ablation isolates these components. Historical runs show operational differences between writing modes but cannot attribute final quality gains (Appendix~\ref{app:archive}). Archived Opus prompts match current inputs, while exact historical input equality remains unverified. Generation and judge logs lack a complete version-linked provenance chain.

The final harness decks benefited from repeated engineering changes driven partly by Opus feedback. Direct-generation baselines were single samples without comparable adaptation budgets or reserved test tasks. Changes in serving and reasoning effort confound the stronger 27B prompt comparison. Separate-session comparisons also confound system differences with session offsets. The text-only comparisons control template markers and presentation only partially, leave length differences, and may introduce within-session contrast effects.

Development can adapt to its judge \citep{manheim2018goodhart,gao2023scaling}. Gains observed by other model families suggest they are not specific to Opus, but shared preferences remain possible. The text-only passes each provide one assessment per judge under different preprocessing and order. They are not identical-condition replications.

\section{Reproducibility}\label{sec:repro}

% NUMBERS-FROM: data/counts.json; data/audit_checks.json.
The development table contains 187 records and the panel table 1,479. Anonymised scores, analysis scripts and an exclusion register are available from the authors on request. The archived API expenditure was 8.2 US dollars for judging and strong-prompt 27B generation. This excludes local compute, agent sessions and development labour. Decks, verdict texts, generator inputs and the parser remain private because they contain bank and issuer names.

\section{Conclusion}

% NUMBERS-FROM: data/textonly.json (blind); data/audit_checks.json.
On the development cases, the complete harness receives consistently higher scores than direct generation by its 27B model. Against the short prompt, the cleaned text-only gains are 20.4 to 33.6 points. Its scores are close to direct Opus generation in this setting. Judges agree on broad development progress but less on the ranking of final decks. A change of judge or session can also shift scores without changing the deck. Comparisons should therefore hold the judge configuration fixed and repeat grading before treating small differences as progress (Appendix~\ref{app:protocol}).

\bibliography{refs}

\begin{thebibliography}{39}
\providecommand{\natexlab}[1]{#1}

\bibitem[{Atil et~al.(2024)Atil, Aykent, Chittams, Fu, Passonneau, Radcliffe,
  Rajagopal, Sloan, Tudrej, Ture, Wu, Xu, and Baldwin}]{atil2024nondeterminism}
Berk Atil, Sarp Aykent, Alexa Chittams, Lisheng Fu, Rebecca~J. Passonneau, Evan
  Radcliffe, Guru~Rajan Rajagopal, Adam Sloan, Tomasz Tudrej, Ferhan Ture, Zhe
  Wu, Lixinyu Xu, and Breck Baldwin. 2024.
\newblock Non-determinism of ``deterministic'' {LLM} settings.
\newblock \emph{arXiv preprint arXiv:2408.04667}.

\bibitem[{Bavaresco et~al.(2025)Bavaresco, Bernardi, Bertolazzi, Elliott,
  Fern{\'a}ndez, Gatt, Ghaleb, Giulianelli, Hanna, Koller, Martins, Mondorf,
  Neplenbroek, Pezzelle, Plank, Schlangen, Suglia, Surikuchi, Takmaz, and
  Testoni}]{bavaresco2024llms}
Anna Bavaresco, Raffaella Bernardi, Leonardo Bertolazzi, Desmond Elliott,
  Raquel Fern{\'a}ndez, Albert Gatt, Esam Ghaleb, Mario Giulianelli, Michael
  Hanna, Alexander Koller, Andr{\'e} F.~T. Martins, Philipp Mondorf, Vera
  Neplenbroek, Sandro Pezzelle, Barbara Plank, David Schlangen, Alessandro
  Suglia, Aditya~K. Surikuchi, Ece Takmaz, and Alberto Testoni. 2025.
\newblock {LLMs} instead of human judges? {A} large scale empirical study
  across 20 {NLP} evaluation tasks.
\newblock In \emph{Proceedings of the 63rd Annual Meeting of the Association
  for Computational Linguistics (Volume 2: Short Papers)}, pages 238--255.

\bibitem[{Bland and Altman(1986)}]{bland1986statistical}
J.~Martin Bland and Douglas~G. Altman. 1986.
\newblock Statistical methods for assessing agreement between two methods of
  clinical measurement.
\newblock \emph{The Lancet}, 327(8476):307--310.

\bibitem[{Card et~al.(2020)Card, Henderson, Khandelwal, Jia, Mahowald, and
  Jurafsky}]{card2020power}
Dallas Card, Peter Henderson, Urvashi Khandelwal, Robin Jia, Kyle Mahowald, and
  Dan Jurafsky. 2020.
\newblock With little power comes great responsibility.
\newblock In \emph{Proceedings of the 2020 Conference on Empirical Methods in
  Natural Language Processing}, pages 9263--9274.

\bibitem[{Chen et~al.(2021)Chen, Chen, Smiley, Shah, Borova, Langdon, Moussa,
  Beane, Huang, Routledge, and Wang}]{chen2021finqa}
Zhiyu Chen, Wenhu Chen, Charese Smiley, Sameena Shah, Iana Borova, Dylan
  Langdon, Reema Moussa, Matt Beane, Ting-Hao Huang, Bryan Routledge, and
  William~Yang Wang. 2021.
\newblock {FinQA}: A dataset of numerical reasoning over financial data.
\newblock In \emph{Proceedings of the 2021 Conference on Empirical Methods in
  Natural Language Processing}, pages 3697--3711.

\bibitem[{Chiang and Lee(2023)}]{chiang2023can}
Cheng-Han Chiang and Hung-yi Lee. 2023.
\newblock Can large language models be an alternative to human evaluations?
\newblock In \emph{Proceedings of the 61st Annual Meeting of the Association
  for Computational Linguistics (Volume 1: Long Papers)}, pages 15607--15631.

\bibitem[{Cohen(1960)}]{cohen1960coefficient}
Jacob Cohen. 1960.
\newblock A coefficient of agreement for nominal scales.
\newblock \emph{Educational and Psychological Measurement}, 20(1):37--46.

\bibitem[{Cohen(1968)}]{cohen1968weighted}
Jacob Cohen. 1968.
\newblock Weighted kappa: Nominal scale agreement provision for scaled
  disagreement or partial credit.
\newblock \emph{Psychological Bulletin}, 70(4):213--220.

\bibitem[{Dror et~al.(2018)Dror, Baumer, Shlomov, and
  Reichart}]{dror2018hitchhiker}
Rotem Dror, Gili Baumer, Segev Shlomov, and Roi Reichart. 2018.
\newblock The hitchhiker's guide to testing statistical significance in natural
  language processing.
\newblock In \emph{Proceedings of the 56th Annual Meeting of the Association
  for Computational Linguistics (Volume 1: Long Papers)}, pages 1383--1392.

\bibitem[{Dubois et~al.(2024)Dubois, Galambosi, Liang, and
  Hashimoto}]{dubois2024length}
Yann Dubois, Bal{\'a}zs Galambosi, Percy Liang, and Tatsunori~B. Hashimoto.
  2024.
\newblock Length-controlled {AlpacaEval}: A simple way to debias automatic
  evaluators.
\newblock \emph{arXiv preprint arXiv:2404.04475}.

\bibitem[{Fu et~al.(2022)Fu, Wang, McDuff, and Song}]{fu2022doc2ppt}
Tsu-Jui Fu, William~Yang Wang, Daniel McDuff, and Yale Song. 2022.
\newblock {DOC2PPT}: Automatic presentation slides generation from scientific
  documents.
\newblock In \emph{Proceedings of the AAAI Conference on Artificial
  Intelligence}, volume~36, pages 634--642.

\bibitem[{Gandhi et~al.(2026)Gandhi, Suryanarayanan, Anwar, Shaik, Desai,
  Nguyen, Raza, Chowdhary, and Neubig}]{gandhi2026ppteval}
Apurva Gandhi, Vishwas Suryanarayanan, Raja~Hasnain Anwar, Firoz Shaik,
  Shubhang Desai, Thong~Q. Nguyen, Muhammad~Taqi Raza, Vishal Chowdhary, and
  Graham Neubig. 2026.
\newblock \href {https://proceedings.mlr.press/v306/gandhi26a.html}
  {{PPT-Eval}: A benchmark for computer-use agents on {PowerPoint} tasks}.
\newblock In \emph{Proceedings of the 43rd International Conference on Machine
  Learning}, volume 306 of \emph{Proceedings of Machine Learning Research},
  pages 32917--32957. PMLR.

\bibitem[{Gao et~al.(2023)Gao, Schulman, and Hilton}]{gao2023scaling}
Leo Gao, John Schulman, and Jacob Hilton. 2023.
\newblock Scaling laws for reward model overoptimization.
\newblock In \emph{Proceedings of the 40th International Conference on Machine
  Learning}, volume 202 of \emph{PMLR}, pages 10835--10866.

\bibitem[{Gu et~al.(2024)Gu, Jiang, Shi, Tan, Zhai, Xu, Li, Shen, Ma, Liu,
  Wang, Zhang, Wang, Gao, Ni, and Guo}]{gu2024survey}
Jiawei Gu, Xuhui Jiang, Zhichao Shi, Hexiang Tan, Xuehao Zhai, Chengjin Xu, Wei
  Li, Yinghan Shen, Shengjie Ma, Honghao Liu, Saizhuo Wang, Kun Zhang, Yuanzhuo
  Wang, Wen Gao, Lionel Ni, and Jian Guo. 2024.
\newblock A survey on {LLM}-as-a-judge.
\newblock \emph{arXiv preprint arXiv:2411.15594}.

\bibitem[{Haldar and Hockenmaier(2025)}]{haldar2025rating}
Rajarshi Haldar and Julia Hockenmaier. 2025.
\newblock Rating roulette: Self-inconsistency in {LLM}-as-a-judge frameworks.
\newblock In \emph{Findings of the Association for Computational Linguistics:
  EMNLP 2025}, pages 24986--25004.

\bibitem[{Islam et~al.(2023)Islam, Kannappan, Kiela, Qian, Scherrer, and
  Vidgen}]{islam2023financebench}
Pranab Islam, Anand Kannappan, Douwe Kiela, Rebecca Qian, Nino Scherrer, and
  Bertie Vidgen. 2023.
\newblock {FinanceBench}: A new benchmark for financial question answering.
\newblock \emph{arXiv preprint arXiv:2311.11944}.

\bibitem[{Kim et~al.(2024)Kim, Shin, Cho, Jang, Longpre, Lee, Yun, Shin, Kim,
  Thorne, and Seo}]{kim2024prometheus}
Seungone Kim, Jamin Shin, Yejin Cho, Joel Jang, Shayne Longpre, Hwaran Lee,
  Sangdoo Yun, Seongjin Shin, Sungdong Kim, James Thorne, and Minjoon Seo.
  2024.
\newblock Prometheus: Inducing fine-grained evaluation capability in language
  models.
\newblock In \emph{International Conference on Learning Representations}.

\bibitem[{Koo and Li(2016)}]{koo2016guideline}
Terry~K. Koo and Mae~Y. Li. 2016.
\newblock A guideline of selecting and reporting intraclass correlation
  coefficients for reliability research.
\newblock \emph{Journal of Chiropractic Medicine}, 15(2):155--163.

\bibitem[{Li(2026)}]{li2026whodrifted}
Yitao Li. 2026.
\newblock \href {https://arxiv.org/abs/2606.15474v1} {Who drifted: the system
  or the judge? anytime-valid attribution in {LLM} evaluation pipelines}.
\newblock \emph{arXiv preprint arXiv:2606.15474}.

\bibitem[{Liu et~al.(2023)Liu, Iter, Xu, Wang, Xu, and Zhu}]{liu2023geval}
Yang Liu, Dan Iter, Yichong Xu, Shuohang Wang, Ruochen Xu, and Chenguang Zhu.
  2023.
\newblock {G-Eval}: {NLG} evaluation using {GPT-4} with better human alignment.
\newblock In \emph{Proceedings of the 2023 Conference on Empirical Methods in
  Natural Language Processing}, pages 2511--2522.

\bibitem[{Madaan et~al.(2024)Madaan, Singh, Schaeffer, Poulton, Koyejo,
  Stenetorp, Narang, and Hupkes}]{madaan2024variance}
Lovish Madaan, Aaditya~K. Singh, Rylan Schaeffer, Andrew Poulton, Sanmi Koyejo,
  Pontus Stenetorp, Sharan Narang, and Dieuwke Hupkes. 2024.
\newblock Quantifying variance in evaluation benchmarks.
\newblock \emph{arXiv preprint arXiv:2406.10229}.

\bibitem[{Manheim and Garrabrant(2018)}]{manheim2018goodhart}
David Manheim and Scott Garrabrant. 2018.
\newblock Categorizing variants of {Goodhart's} law.
\newblock \emph{arXiv preprint arXiv:1803.04585}.

\bibitem[{McGraw and Wong(1996)}]{mcgraw1996forming}
Kenneth~O. McGraw and Seok~P. Wong. 1996.
\newblock Forming inferences about some intraclass correlation coefficients.
\newblock \emph{Psychological Methods}, 1(1):30--46.

\bibitem[{Miller(2024)}]{miller2024errorbars}
Evan Miller. 2024.
\newblock Adding error bars to evals: A statistical approach to language model
  evaluations.
\newblock \emph{arXiv preprint arXiv:2411.00640}.

\bibitem[{Panickssery et~al.(2024)Panickssery, Bowman, and
  Feng}]{panickssery2024self}
Arjun Panickssery, Samuel~R. Bowman, and Shi Feng. 2024.
\newblock {LLM} evaluators recognize and favor their own generations.
\newblock In \emph{Advances in Neural Information Processing Systems 37}.

\bibitem[{Schroeder and Wood-Doughty(2024)}]{schroeder2024trust}
Kayla Schroeder and Zach Wood-Doughty. 2024.
\newblock Can you trust {LLM} judgments? {R}eliability of {LLM}-as-a-judge.
\newblock \emph{arXiv preprint arXiv:2412.12509}.

\bibitem[{Shrout and Fleiss(1979)}]{shrout1979intraclass}
Patrick~E. Shrout and Joseph~L. Fleiss. 1979.
\newblock Intraclass correlations: Uses in assessing rater reliability.
\newblock \emph{Psychological Bulletin}, 86(2):420--428.

\bibitem[{Thakur et~al.(2025)Thakur, Choudhary, Ramayapally, Vaidyanathan, and
  Hupkes}]{thakur2024judging}
Aman~Singh Thakur, Kartik Choudhary, Venkat~Srinik Ramayapally, Sankaran
  Vaidyanathan, and Dieuwke Hupkes. 2025.
\newblock Judging the judges: Evaluating alignment and vulnerabilities in
  {LLMs}-as-judges.
\newblock In \emph{Proceedings of the Fourth Workshop on Generation, Evaluation
  and Metrics (GEM)}, pages 404--430.

\bibitem[{Wang et~al.(2024)Wang, Li, Chen, Cai, Zhu, Lin, Cao, Kong, Liu, Liu,
  and Sui}]{wang2024fair}
Peiyi Wang, Lei Li, Liang Chen, Zefan Cai, Dawei Zhu, Binghuai Lin, Yunbo Cao,
  Lingpeng Kong, Qi~Liu, Tianyu Liu, and Zhifang Sui. 2024.
\newblock Large language models are not fair evaluators.
\newblock In \emph{Proceedings of the 62nd Annual Meeting of the Association
  for Computational Linguistics (Volume 1: Long Papers)}, pages 9440--9450.

\bibitem[{Wang et~al.(2026)Wang, Zhu, Hu, Yuan, Chen, Senthil, Hajishirzi,
  Tsvetkov, Dasigi, and Xiao}]{wang2026harnessevolution}
Yike Wang, Huaisheng Zhu, Zhengyu Hu, Yige Yuan, Zhengyu Chen, Shakti Senthil,
  Hannaneh Hajishirzi, Yulia Tsvetkov, Pradeep Dasigi, and Teng Xiao. 2026.
\newblock \href {https://arxiv.org/abs/2607.12227v2} {Rethinking the evaluation
  of harness evolution for agents}.
\newblock \emph{arXiv preprint arXiv:2607.12227}.

\bibitem[{Weir(2005)}]{weir2005quantifying}
Joseph~P. Weir. 2005.
\newblock Quantifying test-retest reliability using the intraclass correlation
  coefficient and the {SEM}.
\newblock \emph{Journal of Strength and Conditioning Research}, 19(1):231--240.

\bibitem[{Wilcoxon(1945)}]{wilcoxon1945individual}
Frank Wilcoxon. 1945.
\newblock Individual comparisons by ranking methods.
\newblock \emph{Biometrics Bulletin}, 1(6):80--83.

\bibitem[{Xie et~al.(2024)Xie, Han, Chen, Xiang, Zhang et~al.}]{xie2024finben}
Qianqian Xie, Weiguang Han, Zhengyu Chen, Ruoyu Xiang, Xiao Zhang, et~al. 2024.
\newblock {FinBen}: A holistic financial benchmark for large language models.
\newblock In \emph{Advances in Neural Information Processing Systems 37,
  Datasets and Benchmarks Track}.

\bibitem[{Ye et~al.(2025)Ye, Wang, Huang, Chen, Zhang, Moniz, Gao, Geyer,
  Huang, Chen, Chawla, and Zhang}]{ye2024justice}
Jiayi Ye, Yanbo Wang, Yue Huang, Dongping Chen, Qihui Zhang, Nuno Moniz, Tian
  Gao, Werner Geyer, Chao Huang, Pin-Yu Chen, Nitesh~V. Chawla, and Xiangliang
  Zhang. 2025.
\newblock Justice or prejudice? {Q}uantifying biases in {LLM}-as-a-judge.
\newblock In \emph{International Conference on Learning Representations}.

\bibitem[{Yen et~al.(2026)Yen, Poeltl, Gear, Meng, Fan, Shen, Liu, Bauyrzhan,
  Shea, Du, Liu, Guetta, and Namkoong}]{yen2026mbabench}
Thomson Yen, Julian Poeltl, Harshith~Srinivas Gear, Yilin Meng, Joshua Fan,
  Adam Shen, Yili Liu, Ali Bauyrzhan, Patrick Shea, Siri Du, Haoyang Liu,
  Daniel Guetta, and Hongseok Namkoong. 2026.
\newblock \href {https://arxiv.org/abs/2605.22664v5} {{MBABench}: Evaluating
  {LLM} agents on end-to-end spreadsheet tasks in finance}.
\newblock \emph{arXiv preprint arXiv:2605.22664}.

\bibitem[{Zeng et~al.(2024)Zeng, Yu, Gao, Meng, Goyal, and
  Chen}]{zeng2024llmbar}
Zhiyuan Zeng, Jiatong Yu, Tianyu Gao, Yu~Meng, Tanya Goyal, and Danqi Chen.
  2024.
\newblock Evaluating large language models at evaluating instruction following.
\newblock In \emph{International Conference on Learning Representations}.

\bibitem[{Zheng et~al.(2025)Zheng, Guan, Kong, Zhang, Zheng, Zhou, Lin, Lu,
  Han, and Sun}]{zheng2025pptagent}
Hao Zheng, Xinyan Guan, Hao Kong, Wenkai Zhang, Jia Zheng, Weixiang Zhou,
  Hongyu Lin, Yaojie Lu, Xianpei Han, and Le~Sun. 2025.
\newblock \href {https://doi.org/10.18653/v1/2025.emnlp-main.728} {{PPTAgent}:
  Generating and evaluating presentations beyond text-to-slides}.
\newblock In \emph{Proceedings of the 2025 Conference on Empirical Methods in
  Natural Language Processing}, pages 14402--14418.

\bibitem[{Zheng et~al.(2023)Zheng, Chiang, Sheng, Zhuang, Wu, Zhuang, Lin, Li,
  Li, Xing, Zhang, Gonzalez, and Stoica}]{zheng2023judging}
Lianmin Zheng, Wei-Lin Chiang, Ying Sheng, Siyuan Zhuang, Zhanghao Wu, Yonghao
  Zhuang, Zi~Lin, Zhuohan Li, Dacheng Li, Eric~P. Xing, Hao Zhang, Joseph~E.
  Gonzalez, and Ion Stoica. 2023.
\newblock Judging {LLM}-as-a-judge with {MT-Bench} and {Chatbot Arena}.
\newblock In \emph{Advances in Neural Information Processing Systems 36,
  Datasets and Benchmarks Track}.

\bibitem[{Zhu et~al.(2023)Zhu, Wang, and Wang}]{zhu2023judgelm}
Lianghui Zhu, Xinggang Wang, and Xinlong Wang. 2023.
\newblock {JudgeLM}: Fine-tuned large language models are scalable judges.
\newblock \emph{arXiv preprint arXiv:2310.17631}.

\end{thebibliography}

\appendix
\raggedbottom

\section{The original pipeline}\label{app:pipeline}

The original pipeline targets debt capital markets and coverage pitches. Each stage uses a separate model session with specified input files and versioned outputs.

Collection runs in two passes. A first web collection fills a manifest of required fields and is checked for inconsistencies and grey areas; a second collection targets the gaps and conflicts that check found and resolves each conflict by its cause. The result is a self-contained data block and a data contract that fixes what later steps may no longer guess. A calculation engine performs arithmetic from that block and writes a reconciliation file with one key per number and an audit log. The model transcribes inputs and adds options, a recommended sequence and stress cases.

Three documents are drafted from the same reconciliation file in separate conversations, for the executive committee, the risk committee and the client, each citing numbers by key. Senior-editing passes then work on tone, consistency and finish without inventing or recomputing anything. A script checks every cited number against the reconciliation file (discrepancies, orphan keys, numbers without a key, forbidden tokens) and blocks the process on any failure. An adversarial review in a fresh context, run on a model at least as strong as the producing one, then judges scope, storyline and the stress cases and returns a verdict with a routing decision. A failure is corrected at its root step, never on the slide, and all documents are regenerated before the review is run again. Only a passing document is laid out, with a blocking visual check per slide.

A defect register is the one file that crosses runs. Each review reads it first and appends findings. Periodic reviews turn recurring error classes into rules, preferably implemented in code, and retire rules that no longer detect defects. The design calls for a mandatory calculation engine, blocking verification, shorter prompts split into more steps, structured output and internal data sources. It also proposes a strong review model and seeded defects to validate checks before production.

\section{Harness gap by criterion and on text alone}\label{app:crit}

Table~\ref{tab:harness-crit} breaks the harness advantage of Section~\ref{sec:res-harness} down by rubric criterion, and Table~\ref{tab:textonly} gives the cleaned text-only pass.

% NUMBERS-FROM: data/textonly.json (first).
The first text-only pass retained identifying template material. Its mean harness advantage ranged from 17.5 to 36.4 points against the direct 27B and from $-4.1$ to $+1.9$ against direct Opus. Each judge preferred the harness to the 27B on sixteen or seventeen deliverables. These earlier results are secondary to the cleaned pass.

\begin{table}[!t]
\centering\scriptsize\setlength{\tabcolsep}{3pt}
\begin{tabular}{lrrr}
\toprule
Criterion (weight) & 27B short (7) & 27B strong (5) & Opus short (7) \\
\midrule
Reader's problem (15) & +7\% & +10\% & 0\% \\
Synthesis (15) & +8\% & +14\% & +1\% \\
Recommendation (15) & +6\% & +6\% & 0\% \\
Depth, scenarios (15) & +15\% & +13\% & $-$1\% \\
Consistency (20) & +19\% & +32\% & +3\% \\
Storyline (10) & +8\% & +13\% & +1\% \\
Why this bank (5) & +23\% & +3\% & +8\% \\
Visual (5) & +20\% & +38\% & +19\% \\
\bottomrule
\end{tabular}

\caption{Mean advantage of the harness deck (27B, round 7) over each no-harness condition, per criterion, as a share of the criterion's weight, averaged over the judges that graded the condition (number of judges in the column header). Short and strong refer to the prompt.}
\label{tab:harness-crit}
\end{table}

\begin{table}[!t]
\centering\scriptsize\setlength{\tabcolsep}{3pt}
\begin{tabular}{lccc}
\toprule
 & \multicolumn{2}{c}{Harness minus no harness} & Length \\
\cmidrule(lr){2-3}
Judge & 27B, short & Opus, short & $\rho$, Opus \\
\midrule
Claude Sonnet & \begin{tabular}[t]{@{}c@{}}+33.6 (17/17)\\ {[}30.1, 37.2{]}\end{tabular} & \begin{tabular}[t]{@{}c@{}}+0.4 (10/17)\\ {[}$-$3.4, 3.6{]}\end{tabular} & 0.46 \\[2pt]
Claude Opus & \begin{tabular}[t]{@{}c@{}}+29.8 (17/17)\\ {[}24.8, 34.4{]}\end{tabular} & \begin{tabular}[t]{@{}c@{}}$-$4.7 (4/17)\\ {[}$-$8.2, $-$1.4{]}\end{tabular} & 0.54 \\[2pt]
DeepSeek V4.1 Flash & \begin{tabular}[t]{@{}c@{}}+23.8 (17/17)\\ {[}19.3, 28.3{]}\end{tabular} & \begin{tabular}[t]{@{}c@{}}$-$1.1 (6/17)\\ {[}$-$3.9, 1.1{]}\end{tabular} & 0.07 \\[2pt]
GPT-6 Luna Pro & \begin{tabular}[t]{@{}c@{}}+23.2 (17/17)\\ {[}18.1, 28.6{]}\end{tabular} & \begin{tabular}[t]{@{}c@{}}$-$2.1 (8/17)\\ {[}$-$4.6, 0.4{]}\end{tabular} & 0.31 \\[2pt]
GLM 5.3 FlashX & \begin{tabular}[t]{@{}c@{}}+20.4 (17/17)\\ {[}17.0, 24.0{]}\end{tabular} & \begin{tabular}[t]{@{}c@{}}+0.8 (11/17)\\ {[}$-$1.5, 2.8{]}\end{tabular} & 0.02 \\[2pt]
\bottomrule
\end{tabular}

\caption{Harness text (27B, round 7) minus text written without the harness in the cleaned text-only pass, out of 95 without visual quality. Cells give 95\% bootstrap intervals and the number of deliverables where the harness text scores higher. All three versions of a deliverable were graded in the same session. The last column is the Spearman correlation, over deliverables, between the harness margin over the Opus text and the difference in word count. All totals and criterion scores are available in this pass.}
\label{tab:textonly}
\end{table}

\FloatBarrier

\section{Protocol for grading with an LLM judge}\label{app:protocol}

\begin{enumerate}\setlength{\itemsep}{1pt}
\item Repeat grading with decks assigned to sessions at random. Two passes expose discrepancies; estimating variability precisely requires a larger design matched to the desired precision.
\item Treat decks graded in the same session as a cluster when testing a change.
\item Estimate per-document and campaign-level variability, report its uncertainty and avoid treating a small change as conclusive from one pass.
\item Compare systems within a single judge and configuration, and anchor any change of judge on a shared set of decks.
\item Read the continuous score rather than a ship-or-not verdict when the score lies near the verdict threshold.
\item Record model, provider, date, reasoning effort and input format for every judge call.
\item When systems differ in presentation, add a pass on the text alone and check whether document length tracks the score.
\item Check a subset of decks with human experts before trusting absolute levels.
\end{enumerate}

\section{Judge configurations}\label{app:config}

\begin{table}[H]
\centering\footnotesize\setlength{\tabcolsep}{4pt}
\begin{tabular}{llll}
\toprule
Judge & Access & Effort & Slides \\
\midrule
Opus & agent & default & full size \\
Sonnet & agent & default & full size \\
Haiku & agent & default & full size \\
DeepSeek V4.1 Flash & API & low & 1600 px \\
GLM 5.3 FlashX & API & low & 1600 px \\
GPT-6 Luna Pro & API & low & 1600 px \\
Gemini 3.8 Flash & API & low & 1600 px \\
Qwen3.8-27B & on premises & not set & 1600 px \\
\bottomrule
\end{tabular}
\caption{Judge configurations. Effort refers to reasoning; sampling settings are provider defaults. All judges received extracted text. Text-only passes omitted slide images (Section~\ref{sec:setting}).}
\label{tab:config}
\end{table}

\section{Judge instruction}\label{app:instruction}

We translate and reformat the French instruction.

\begin{quote}\small
You are an investment-bank managing director. Your area is debt capital markets, credit or M\&A, depending on the deliverable. You grade draft presentations, given as rendered images (four slides per image) and extracted text. You do not know who made them or which are recent. Read the slides first and the text second, to quote exactly.

Use the following weighted rubric. Criteria (1) reader orientation, (2) ten-second synthesis, (3) recommendation and ask (what, when, who), and (4) financial depth and scenarios each receive 15 points. Criterion (5) consistency and auditability receives 20. Criterion (6) title storyline receives 10. Criteria (7) why this bank and (8) visual quality and meeting effectiveness each receive 5.

For each deck, give criterion scores with justifications and exact quotations with page numbers, then a total out of 100. Answer whether you would send it as is (yes / after one round of corrections / no), with the reason. List the five most serious defects with quotations, pages and precise corrections. Be cold and exact. A claim without a quotation does not count. Check the arithmetic you can redo.

``Yes'' means you would send it to the client or the committee as it is, without retouching. A figure unverifiable from the pages or a contradiction between pages prevents a ``yes''. So does machine-sounding wording or a layout defect a director would notice.
\end{quote}

\FloatBarrier
\section{Verdicts against scores}\label{app:verdict}

\begin{figure}[H]
\centering
\includegraphics[width=\columnwidth]{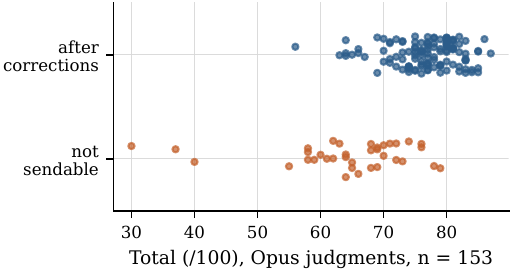}
\caption{Total of every development Opus judgment by verdict ($n=153$, vertical jitter added).}
\label{fig:verdict}
\end{figure}

% NUMBERS-FROM: data/results.json (verdict diagnostics and repeated pairs).
A threshold at 70 classifies 86\% of development Opus judgments, but the two verdict classes overlap between 56 and 79 (Figure~\ref{fig:verdict}). The same threshold, chosen on these data, would have flipped one and two decks between the two Opus passes of rounds 7 and 8. The development Sonnet pass returned ``send as is'' 7 times in 34, always at 92 or above.

\FloatBarrier
\section{Paired scores on identical decks}\label{app:pairs}

\begin{figure}[t]
\centering
\includegraphics[width=0.62\columnwidth]{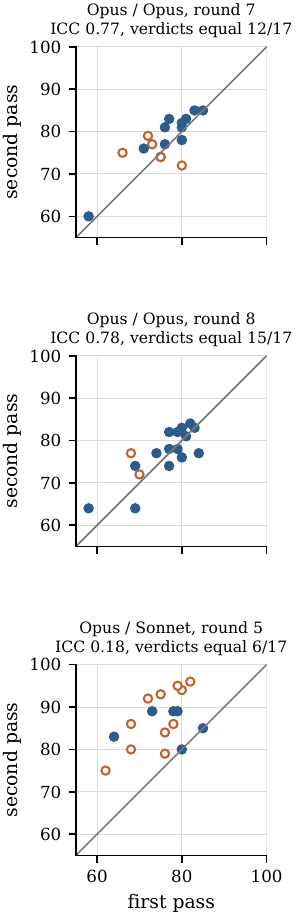}
\caption{Per-deck totals from two passes on identical decks: Opus/Opus in rounds 7 and 8 (top, middle) and Opus/Sonnet on 24 September in round 5 (bottom). Open points changed verdict between the passes.}
\label{fig:pairs}
\end{figure}

Figure~\ref{fig:pairs} plots, deck by deck, the scores of the three pairs of passes summarised in Table~\ref{tab:reliability}.

\section{Judge agreement}\label{app:agree}

\begin{figure}[t]
\centering
\includegraphics[width=\columnwidth]{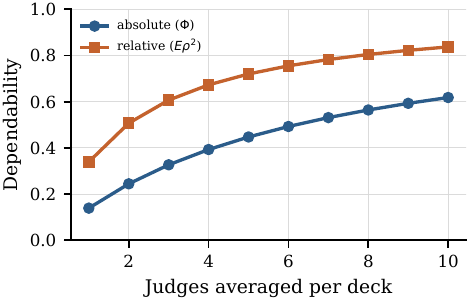}
\caption{Projected dependability when averaging $n$ judges, conditional on the round-7 variance decomposition. Curves show absolute ($\Phi$) and relative ($E\rho^2$) coefficients. The residual includes interaction and error; extrapolation beyond this convenience panel is unvalidated.}
\label{fig:dstudy}
\end{figure}

Figure~\ref{fig:dstudy} projects dependability when averaging judges. Appendix~\ref{app:variance} gives the model and its assumptions. Round-specific rank agreement is shown in Figure~\ref{fig:round-agreement}.

\section{Evaluation inventory and archives}\label{app:archive}

% NUMBERS-FROM: data/release_summary.json; data/counts.json.
Table~\ref{tab:inventory} separates requested evaluation records from recoverable scores. These are repeated measurements of the same deliverables, not independent documents.

\begin{table}[H]
\centering\scriptsize\setlength{\tabcolsep}{5pt}
\begin{tabular}{lrr}
\toprule
Phase & Records & Scores \\
\midrule
Development & 187 & 187 \\
Panel development decks & 493 & 491 \\
Short-prompt comparisons & 238 & 238 \\
Strong-prompt comparisons & 238 & 236 \\
First text-only pass & 255 & 254 \\
Cleaned text-only pass & 255 & 255 \\
\midrule
Total & 1,666 & 1,661 \\
\bottomrule
\end{tabular}

\caption{Records and recoverable totals by evaluation phase. The strong-prompt phase includes the harness decks regraded in mixed sessions. Missing totals remain missing; none are imputed.}
\label{tab:inventory}
\end{table}

% NUMBERS-FROM: data/counts.json; data/audit_checks.json.
The archive contains 1,666 evaluation records, of which 1,661 contain recoverable totals. A parser preserves fractional scores and recognises both total and score headings. Development criterion sums match totals in 185 of 187 records. All controlled Opus and Sonnet records contain complete criteria, including both text-only passes. API and on-premises judges omit totals in four of 918 records. Haiku lacks one total and ten categorical verdicts. Parsed scores were checked against the raw judge responses.

% NUMBERS-FROM: data/provenance_audit.json.
An archive audit identified five pilot sessions containing 21 scores outside the main collection, three earlier verdicts using different deck sets or formats, and a truncated test response. Earlier calibration archives contain 25 reports with 50 presentation-order rows. Their grids, candidates and admissibility rules differ from the main study. We retain them in an exclusion register rather than pooling them with the main scores. The pilot exclusion rationale was not fully recorded contemporaneously, limiting retrospective selection checks.

% NUMBERS-FROM: data/provenance_audit.json.
One earlier case compared free-field writing with compiled assertions. The archived diagnostics record 121 versus 41 model calls and approximately 191 versus 85 seconds. Repetition warnings increased from six to 30, and blocking defects from zero to one. Page titles and exhibits match, but prose and the thesis differ, and the claimed common registry lacks a preserved snapshot. Later tuning reduced calls further; it is not an independent replication. These archives document operational tradeoffs, not a component effect on final-deck quality.

\FloatBarrier
\section{Statistical estimands}\label{app:variance}

Fixed-group repetitions measure within-context variability. Reassigned sessions also vary grouping and order; configuration comparisons vary further factors. Their detectable-change estimates are not interchangeable.

% NUMBERS-FROM: data/results.json (pooled_retest); data/crossfam.json; data/audit_checks.json.
The pooled Opus estimate reuses seventeen deliverables across 34 differences. Deck-bootstrap and chi-square intervals omit repeated-document and session dependence.

Each round uses the first original-grouping score per deck and judge, from development for Opus and the controlled panel for others. Replications are not averaged. In $Y_{dj}=\mu+a_d+b_j+e_{dj}$, $a_d$ and $b_j$ represent deck and judge effects. The residual combines interaction and error. With $D$ decks and $J$ judges, two-way mean squares give
\begin{align}
\hat v_d&=\max\{(MS_d-MS_e)/J,0\},\\
\hat v_j&=\max\{(MS_j-MS_e)/D,0\},\\
\hat v_e&=MS_e.
\end{align}
Negative component estimates are truncated to zero. Session and model-family effects are not fitted. For the projected mean of $k$ judges,
\begin{align}
\Phi_k&=\frac{\hat v_d}{\hat v_d+(\hat v_j+\hat v_e)/k},\\
E\rho^2_k&=\frac{\hat v_d}{\hat v_d+\hat v_e/k}.
\end{align}
Projections assume uncorrelated judge and residual contributions. They are conditional on this convenience panel, without validated precision guarantees for new judges.

\FloatBarrier
\end{document}